\documentclass{article}

\usepackage[authoryear]{natbib}
\usepackage{cite}
\usepackage{wrapfig}
\usepackage{mathrsfs}
\usepackage{times}
\usepackage{enumerate}
\usepackage{color}
\usepackage{graphicx,epsfig,subfigure}
\usepackage{algorithm,algorithmic}
\usepackage{amsmath,amssymb,xspace}
\usepackage{url}
\usepackage{subfigure}
\usepackage{hyperref}

\newcommand{\sbr}[1]{\left[#1\right]}
\newcommand{\rbr}[1]{\left(#1\right)}
\newcommand{\EE}{\mathbb{E}} 
\newcommand{\ie}{{{i.e.,}}\xspace}
\newcommand{\eg}{{{\em e.g.,}}\xspace}
\newcommand{\cmt}[1]{}
\newcommand{\comment}[1]{}

\newcommand{\ours}{{\textsc{DinTucker}}\xspace}
\newcommand{\oursw}{\textsc{DinTucker}\ensuremath{_{\textrm{W}}}\xspace}
\newcommand{\oursu}{\textsc{DinTucker}\ensuremath{_{\textrm{U}}}\xspace}
\newcommand{\oursg}{\textsc{DinTucker}\ensuremath{_{\textrm{G}}}\xspace}
\newcommand{\hadoop}{\textsc{Hadoop}\xspace}
\newcommand{\mapreduce}{\textsc{MapReduce}\xspace}
\newcommand{\map}{\textsc{Map}\xspace}
\newcommand{\mapper}{\textsc{Mapper}\xspace}
\newcommand{\reduce}{\textsc{Reduce}\xspace}
\newcommand{\reducer}{\textsc{Reducer}\xspace}
\newcommand{\inftucker}{{InfTucker}\xspace}
\newcommand{\InfTucker}{{InfTucker}\xspace}
\newcommand{\tucker}[1]{[\![#1]\!]}
\newcommand{\cbr}[1]{\left\{#1\right\}}

\newcommand{\alanc}[1]{}
\newcommand{\expec}[2]{\EE_{{#1}}\sbr{{#2}}}

\newcommand{\email}[1]{\href{mailto:#1}{#1}}

\newcommand{\tr}{{\rm tr}}
\renewcommand{\vec}{{\rm vec}}

\newcommand{\bi}{{\bf i}}

\newcommand{\mhat}{{\overline{m}}}

\renewcommand{\u}{{\bf u}}

\newcommand{\I}{{\bf I}}

\newcommand{\Mcal}{{\mathcal{M}}}
\newcommand{\N}{\mathcal{N}}  

\newcommand{\Tcal}{{\mathcal{T}}}
\newcommand{\U}{{\bf U}}
\newcommand{\Ucal}{{\mathcal{U}}}
\newcommand{\tU}{{\tilde{\U}}}
\newcommand{\tUcal}{{\tilde{\Ucal}}}

\newcommand{\Wcal}{{\mathcal{W}}}

\newcommand{\Xcal}{{\mathcal{X}}}

\newcommand{\Ycal}{{\mathcal{Y}}}

\newcommand{\Zcal}{{\mathcal{Z}}}

\newcommand{\bLambda}{\mathbf{\Lambda}}

\newcommand{\bSigma}{\boldsymbol{\Sigma}}

\newcommand{\bUpsilon}{\boldsymbol{\Upsilon}}

\newcommand{\bmu}{\boldsymbol{\mu}}

\newcommand{\0}{{\bf 0}}

\newcommand{\ben}{\begin{enumerate}}
\newcommand{\een}{\end{enumerate}}

\begin{document}

\title{DinTucker: Scaling up Gaussian process models on multidimensional arrays with billions of elements}
\author{
       Shandian Zhe\\
       Purdue University\\
       \email{szhe@purdue.edu}
       \and
        Yuan Qi \\
        Purdue University \\
        \email{alanqi@purdue.edu}
        \and
        Youngja  Park\\
        IBM T. J. Watson Research Center\\
        \email{young\_parkatus.ibm.com}
        \and
        Ian Molloy\\
		IBM T. J. Watson Research Center\\
		\email{molloyim@us.ibm.com}
		\and
		Suresh Chari\\
		IBM T. J. Watson Research Center\\
		\email{schari@us.ibm.com}
}

\maketitle

\begin{abstract}
Infinite Tucker Decomposition (InfTucker) and random function prior models, as nonparametric Bayesian models on infinite exchangeable arrays, are more powerful models than widely-used multilinear factorization methods including Tucker and PARAFAC decomposition, (partly) due to their capability of modeling nonlinear relationships between array elements.
Despite their great predictive performance and sound theoretical foundations,  they 
cannot handle massive data due to a prohibitively high training time. 
To overcome this limitation, we present \textbf{D}istributed  \textbf{in}finite \textbf{Tucker} (\ours), a large-scale nonlinear tensor decomposition algorithm on \mapreduce. While maintaining the predictive accuracy of InfTucker, it is scalable on massive data. \ours is based on a new
hierarchical Bayesian model that enables local training of InfTucker on subarrays and information integration from all local training results. We use distributed stochastic gradient descent, coupled with variational inference, to train this model.
We apply \ours to multidimensional arrays with billions of elements from applications in the "Read the Web" project \citep{carlson2010toward} and in information security
and compare it with  the state-of-the-art large-scale tensor decomposition method, GigaTensor.
On both datasets, 
\ours achieves significantly higher prediction accuracy with less computational time.
\end{abstract}

\section{Introduction}

Many real-world datasets with multiple aspects can be described by multidimensional arrays (\ie tensors). For example, an access log database can be represented by an array with three modes
({\em user, file, action}),
patient drug responses by an array with four modes ({\em person, medicine, biomarker, time}), and predicates in knowledge bases by an array with three modes ({\em subject, verb, object}).
Given tensor-valued data, we want to model complex interactions embedded in data (\eg drug interactions) and predict missing elements (\eg unknown drug responses).

InfTucker \citep{XuYQ12} and its generalization, random function prior models \citep{LloydOGR12randomgraph}, are elegant nonparametric Bayesian models, which assign Bayesian priors on multidimensional random arrays with {\em infinite} number of columns for each mode. For the two dimensional case, these arrays are known as doubly infinite row$-$column exchangeable (RCE) arrays \citep{Aldous81,Lauritzen07randomgraph}.
The RCE array, as a generalization of a classical infinite exchangeable sequence, has such a property: its distribution is unchanged when its rows and columns are permuted separately (not necessarily in the same way). The InfTucker model is justified theoretically by the generalization of de Finetti's theorem for the RCE arrays \citep{Aldous81,Lauritzen07randomgraph}. In addition, as shown by \citet{XuYanQi2011}, InfTucker achieves superior predictive performance on several benchmark datasets; compared with previous multidimensional array models, including the Tucker decomposition \citep{Tucker66} and CANDECOMP/PARAFAC (CP) \citep{Harshman70parafac} and their generalizations \citep{Chu09ptucker}, InfTucker leads to an almost three-fold error reduction!

However, a critical bottleneck of \InfTucker and other random function prior models is that they operate on data that can fit in the main memory of a computer. Even with fast approximate inference, their scalability is constrained by the computational power of a single computer. For many applications, the data is easily at the scale of tens of Gigabytes or even Terabytes, making \InfTucker infeasible on a single computer. Although \InfTucker has explored properties of the Kronecker product to reduce the computational cost, it does not employ the power of massive computational parallelism offered by a computer cluster or graphics processing units (GPUs), thus limiting itself to relatively small data.

Recently, \citet{kang2012gigatensor} propose the first distributed PARAFAC decomposition algorithm, GigaTensor, on the \mapreduce framework. For sparse array data it explores sparseness of the nonzero elements in the array and avoids the intermediate data explosion.
The \mapreduce-based  GigaTensor algorithm makes PARAFAC a practical tool for massive array data analysis. However, the  PARAFAC model suffers several limitations: i) as a multilinear model, it cannot capture intricate nonlinear relationships encoded in the data; ii) it cannot handle missing data directly and requires data imputation as a preprocessing step; and iii) it cannot deal with binary or count data in a principled way. Although InfTucker or other random function prior models have limited scalability,
they overcome all the above limitations of the PARAFAC model.

In this paper, we propose  \textbf{D}istributed  \textbf{in}finite  \textbf{Tucker} (\ours), a large-scale nonlinear tensor decomposition algorithm on \mapreduce. It keeps the nonlinear modeling power of InfTucker and other random function prior models and, at the same time, makes Gaussian process (GP) scalable on massive multidimensional array data.
To the best of our knowledge, this paper is the first approach of deploying a GP model in the \mapreduce framework. The main contributions of this paper are the following:
\begin{enumerate}
\item Algorithm. We design a hierarchical Bayesian model that enables local training of InfTucker on subarrays and information integration from all local training results. Based on this model, we develop a distributed inference algorithm based on stochastic gradient descent and implement it using \mapreduce.
\item Scalability. InfTucker decomposes large multidimensional arrays, such as those in Table \ref{tb:large_data} \cmt{\alanc{to be fixed}} with more than $50$ billion elements. This is impossible for previous GP or other random function prior models. Furthermore, \ours enjoys almost linear scalability on the number of computational nodes.
\item Applications. In addition to testing our model on large knowledge bases from the "Read the Web" project \citep{carlson2010toward} \cmt{\alanc{see GigaTensor paper reference 9}}, we apply our model to massive user access log data from a large company, with the goal of detecting potential security threat.  On both datasets, \ours achieves significantly higher prediction accuracy with less computational time using the same \hadoop system.
\end{enumerate}

\cmt{
The rest of the paper is organized as follows. Section 2 presents background information on Tucker decomposition and nonlinear Tucker decomposition by InfTucker. Section 3 describes our proposed \ours model and Section 4 describes the distributed inference algorithm for large multidimensional array analysis. In Section 5, we discuss related works and show experimental results in Section 6.
}

\cmt{
In this section, we present the hierarchial Bayesian model, \ours. We first describe the data and tensor decomposition task in section \ref{sec:td}. Then we briefly introduce InfTucker model in section \ref{sec:inftucker}. Finally, we present \ours based on InfTucker in section \ref{sec:dintucker}.
}
\section{Background}
\subsection{Tensor Decomposition}\label{sec:td}
We denote a  $K$-mode multidimensional array or tensor by $\Mcal \in \mathbb{R}^{m_1 \times m_2 \ldots \times m_K}$, where the $k$-th mode has $m_k$ dimensions. We use $m_{\mathbf{i}}$ ($\bi=(i_1,\ldots, i_K)$) to denote $\Mcal$'s entry at location $\bi$.  Using the vectorization operation, we can stack all of $\Mcal$'s entries in a vector, $\vec(\Mcal)$, with size $\prod_{k=1}^K m_k$ by $1$. In $\vec(\Mcal)$, the entry $\mathbf{i}=(i_1, \ldots, i_K)$ of $\Mcal$ is mapped to the entry at position $j=i_K + \sum_{i=1}^{K-1}(i_k - 1)\prod_{k+1}^K m_k$.

Given a tensor $\mathcal{W} \in \mathbb{R}^{r_1 \times \ldots \times r_K}$ and a matrix $\U \in \mathbb{R}^{s \times r_k}$, a mode-$k$ tensor-matrix multiplication between $\mathcal{W}$ and $\U$ is denoted by $\mathcal{W} \times_k \U$, which is a tensor of size $r_1 \times \ldots \times r_{k-1} \times s \times r_{k+1} \times \ldots \times r_K$. The corresponding entry-wise definition is
\[
(\Wcal \times_k \U)_{i_1\ldots i_{k-1} j i_{k+1}\ldots i_K} = \sum_{i_k=1}^{r_k} w_{i_1\ldots i_K}u_{ji_k}.
\]

The Tucker decomposition of $K$-mode tensor $\Mcal$  is
\[
\Mcal = \Wcal \times_1 \U^{(1)} \times_2 \ldots \times_K \U^{(K)} = \tucker{\Wcal;\U^{(1)},\ldots,\U^{(K)}}\raisetag{.05in}
\]
where $\Wcal \in \mathbb{R}^{r_1 \times \ldots \times r_K}$ is the core tensor, and $\U^{(k)} \in \mathbb{R}^{m_k\times r_k}$ is the $k$-th latent factor matrix.
The tucker decomposition can also be represented in a vectorized form
\[
\vec(\tucker{\Wcal;\U^{(1)},\ldots,\U^{(K)}}) = \U^{(1)} \otimes \ldots \otimes \U^{(K)} \cdot \vec(\Wcal) \raisetag{.05in}
\]
where $\otimes$ is the Kronecker product. If we enforce $r_1=\ldots=r_K$ and restrict the core tensor $\Wcal$ to be diagonal (\ie $W_{i_1\ldots i_K} \neq 0$ only if $i_1=\ldots=i_K$), it reduces to PARAFAC decomposition.
\subsection{Infinite Tucker Decomposition}\label{sec:inftucker}
The infinite Tucker (InfTucker) decomposition  \citep{XuYQ12} generalizes the Tucker decomposition in an infinite feature space based on a tensor-variate GP. The tensor-variate GP is a collection of random variables $\{m(\u^{(1)}, \ldots, \u^{(K)})\}$, $\u^{(k)} \in \mathbb{R}^r$, whose finite joint probability over any set of input locations follows the tensor-variate normal density distribution. Specifically, given $\Ucal = \{\U^{(1)}, \ldots ,\U^{(k)}\}$, the zero mean tensor-variate GP on $\Mcal$ has the probability density function
\begin{align}
&p(\Mcal|\U^{(1)},\ldots,\U^{(K)}) \notag \\
&= \Tcal \N(\Mcal; \0, \Sigma^{(1)},\ldots,\Sigma^{(K)}) \notag \\
&=\N(\vec(\Mcal);\0, \Sigma^{(1)}\otimes \ldots \otimes \Sigma^{(K)}) \notag \\
&=\frac{\exp\cbr{-\frac{1}{2} \|\tucker{\Mcal; (\Sigma^{(1)})^{-\frac{1}{2}},\ldots,(\Sigma^{(K)})^{-\frac{1}{2}}}\|^2}}{(2\pi)^{m/2}\prod_{k=1}^K  |\Sigma^{(k)}|^{-\frac{m}{2m_k}}}\label{eq:tensor-gaussian-proc}
\end{align}
where $m=\prod_k m_k$, $\|\Xcal\|=\sqrt{\sum_{\mathbf{i}}x_\mathbf{i}^2}$, and $\Sigma^{(k)}=k(\U^{(k)},\U^{(k)})$ is the covariance matrix.

The InfTucker model assumes $K$ latent factors $\Ucal = \{\U^{(1)}, \ldots, \U^{(K)}\}$ are sampled from
element-wise Laplace priors $p(\Ucal)$, which encourage sparse estimation for easy model interpretation. Given $\Ucal$, a latent real-valued tensor $\Mcal$ is sampled from the tensor variate Gaussian process, as defined in Equation \eqref{eq:tensor-gaussian-proc}. Then, given $\Mcal$, the observed tensor $\Ycal$ is sampled from a noisy model $p(\Ycal | \Mcal)$. For example, we can use probit models for binary observations and Gaussian models for continuous observations. Thus the joint distribution is
\begin{align}
p(\Ycal, \Mcal, \Ucal) = p(\Ucal)p(\Mcal|\Ucal)p(\Ycal|\Mcal)\label{eq:inftucker_joint}.
\end{align}

By using nonlinear covariance function $k(\u^i,\u^i)$, \InfTucker maps the latent factors in each mode into an infinite feature space and then performs the Tucker decomposition with the core tensor $\Wcal$ of infinite size.
Based on a nonlinear feature mapping,  InfTucker can capture nonlinear relationships between latent factors.

\section{Hierachical Bayesian model for DinTucker}\label{sec:dintucker}

A major bottleneck of InfTucker is that it cannot scale to massive data. It requires the entire data to be stored in the main memory of a single computer; this \mbox{requirement} is not satisfied by many real-world multidimensional array data. Furthermore, InfTucker uses sequential updates and, thus, cannot utilize the massive parallelism offered by a distributed computing environment, such as the \hadoop system.
These limitations stem from a global GP assumption used by InfTucker:  it assumes all entries or elements of the tensor $\Mcal$ are sampled from a global Gaussian process given latent factors $\Ucal$. As a result,  computing the distribution for the global $\Mcal$---$p(\Mcal|\U^{(1)},\ldots, \U^{(K)})$ in Equation \eqref{eq:tensor-gaussian-proc}---requires computing the Kronecker-product of the covariance matrices and its inverse. This matrix inversion is prohibitively expensive. Although \citet{XuYQ12} explore properties of the Kronecker product to avoid naive computation, it still needs to perform eigen-decomposition over the covariance matrix for each mode, which is infeasible for a large dimension $m_k$.
Moreover, all the latent factors are coupled in $p(\Mcal|\U^{(1)},\ldots,\U^{(K)})$ so that we can not distribute the computation over many computational units or conduct online learning.

To overcome these limitations, we propose \ours that assumes the data are sampled from many, smaller \mbox{GP} models, and the latent variables for these GP models are coupled together in a hierarchical Bayesian model. The local GP enables  fast computation over subarrays and the  hierarchical Bayesian model allows information sharing across different subarrays---making distributed inference and online learning possible.

Specifically, we first break the observed multidimensional array $\Ycal$ into $N$
subarrays $\{\Ycal_1, \ldots, \Ycal_N\}$ for multiple computational units (\eg one per \mapper in \mapreduce).
Each subarray is sampled from a GP based on latent factors $\tilde{\Ucal}_n = \{\tU_n^{(1)},\ldots, \\\tU_n^{(K)}\}$. Then we tie these latent factors to the common latent factors  $\Ucal = \{\U^{(1)},\ldots, \\\U^{(K)}\}$ via a prior distribution:
\begin{align}
p(\tilde{\Ucal}_n|\Ucal) &= \prod_{k=1}^K p(\tilde{\U}^{(k)}_n | \U^{(k)}) \notag \\
&= \prod_{k=1}^K \N(\vec(\tilde{\U}^{(k)}_n)|\vec(\U^{(k)}), \lambda\I)
\end{align}
where $\lambda$ is a variance parameter that controls the similarity between $\Ucal$ and $\tilde{\Ucal}_n$.

Furthermore, we use stochastic gradient descent (SGD) to optimize $\{\tilde{\Ucal}_n\}$ and $\Ucal$ due to its computational efficiency and theoretical guarantees. The use of SGD also naturally enables us to deal with dynamic array data with increasing size over time.  To use SGD, we further  break each $\Ycal_n$ into $T_n$ smaller subarrays
$\Ycal_n = \{\Ycal_{n1},\ldots,\Ycal_{n T_n}\}$.
We allow the subarrays from each $\Ycal_n$ to share the same latent factors $\{\tilde{\Ucal}_n\}$. The reason that
we do not need to explicitly introduce another set of latent factors, say, $\{\tilde{\Ucal}_{nt}\}_t$, for subarrays in each $\Ycal_n$ is the following: suppose we have a prior
 $p(\tilde{\Ucal}_{nt}|\tilde{\Ucal}_{n})$ to couple these $\tilde{\Ucal}_{nt}$, we can set $p(\tilde{\Ucal}_{nt}|\tilde{\Ucal}_{n}) = \delta(\tilde{\Ucal}_{nt}-\tilde{\Ucal}_n)$ ($\delta(a)=1$ if and only if $a=0$) without causing conflicts between updates over $\tilde{\Ucal}_{nt}$---since they are updated sequentially. This situation is different from parallel inference over $\tilde{\Ucal}_n$ for which, if we simply set  $\tilde{\Ucal}_n= \Ucal$ for all $n$, we will have conflicts between inconsistent $\tUcal_n$ estimated in parallel from different computational units. The graphical model representation of \ours is shown in Figure \ref{fig:dinftucker}.

Given $\tUcal_n$, a latent real-valued subarray $\Mcal_{nt}$ is sampled from the corresponding local GP. Then we sample the noisy observations $\Ycal_{nt}$ from the latent subarray $\Mcal_{nt}$.
Denoting $\{\Mcal_{nt}\}_{t=1}^{T_n}$ by $\Mcal_n$, we have the joint probability of our model
\begin{align}
&p(\Ucal,\{\tUcal_n, \Mcal_n, \Ycal_n\}_{n=1}^N) \notag \\
&=\prod_{n=1}^N p(\tUcal_n|\Ucal)\prod_{t=1}^{T_n} p(\Mcal_{nt}|\tUcal_n)p(\Ycal_{nt}|\Mcal_{nt}). \raisetag{.45in}
\end{align}
Note that $\Mcal_{nt}$ depends only on its corresponding elements in $\tUcal_n$, instead of the whole $\tUcal_n$, so that the computation of $p(\Mcal_{nt}|\tUcal_n)$ is efficient.

Compared with the joint probability of InfTucker in \eqref{eq:inftucker_joint}, the joint probability of
\ours replaces the global factor $p(\Mcal|\U^{(1)}, \ldots, \U^{(K)})$ (which couples all the latent factors and the whole latent multidimensional array $\Mcal$) by smaller local factors. These local factors require much less memory and processing time than the global factors. More important, the additive nature of these local factors in the log domain enables distributed inference and online learning.

\cmt{
From the modeling perspective, the key difference between the two models is that \ours assume the tensor observations are generated based on multiple tensor-variate Gaussian process where each process is over a subset of latent factors which corresponds to a much smaller subarray, while InfTucker assumes the data are generated based on one tensor-variate Gaussian process over all the latent factors, which corresponds to the whole tensor.
}
{
 \begin{figure}[ht]
\centering
\includegraphics[scale=0.5]{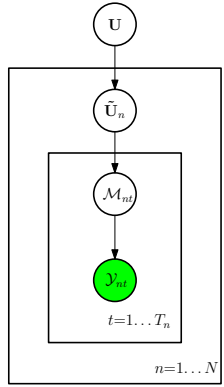}
\caption{The graphical model representation of \ours.}
\label{fig:dinftucker}
\end{figure}
}

\section{Distributed online inference algorithm}
Now we present our distributed online inference \mbox{algorithm} on the \hadoop system. We focus on binary tensor data in this paper, for which we use the probit model for
$p(\Ycal_{nt}|\Mcal_{nt})$. It is straightforward to modify the following presentation to handle continuous and count multidimensional array data.

First, we use data augmentation to decompose the probit model into $p(y_\bi|m_\bi) = \int p(y_\bi|z_\bi)p(z_\bi|m_\bi)dz_\bi$, where
\begin{align}
p(y_\bi|z_\bi) &= \delta(y_\bi=1)\delta(z_\bi>0) + \delta(y_\bi=0)\delta(z_\bi\le 0), \notag \\
p(z_\bi|m_\bi) &= \N(z_\bi|m_\bi,1) \notag
\end{align}
where $\delta(\cdot)$ is the binary indicator function.
For each $\Mcal_{nt} \in \Mcal_n$, we introduce an augmented $\Zcal_{nt}$. Let us denote $\Zcal_n = \{\Zcal_{nt}\}_{t=1}^{T_n}$.  The joint probability of the augmented model is
\begin{align}
&p(\Ucal, \{\tUcal_n, \Mcal_n, \Zcal_n, \Ycal_n\}_{n=1}^N ) \notag \\
=& \prod_{n=1}^N p(\tUcal_n|\Ucal)\prod_{t=1}^{T_n}p(\Mcal_{nt}|\tUcal_n)p(\Zcal_{nt}|\Mcal_{nt})  p(\Ycal_{nt} | \Zcal_{nt}). \raisetag{.45in}
\end{align}

\subsection{Variational approximation}
We then apply variational EM to optimize the \mbox{latent} factors $\Ucal, \{\tUcal_n\}$: in the E-step, we use the variational approximation and, in the M-step, we apply SGD to maximize the variational lower bound over the latent factors.
Specifically, in the E-step, we use a fully factorized distribution $q(\{\Zcal_n, \Mcal_n\}_{n=1}^N) = \prod_{n=1}^N \prod_{t=1}^{T_n} q(\Zcal_{nt})q(\Mcal_{nt})$ to approximate the posterior distribution $p(\{\Zcal_n, \Mcal_n\}_{n=1}^N|\{\Ycal_n, \tUcal_n\}_{n=1}^N, \Ucal)$. The variational inference minimizes the Kullback-Leibler (KL) divergence between the approximate and the exact posteriors  by coordinate descent.
The variational updates for $q(\Zcal_{nt})$ and $q(\Mcal_{nt})$ are the same as those for  $q(\Zcal)$ and  $q(\Mcal)$ in \citep{XuYQ12}.

\cmt{
 listed below for the reference in Section 5 (they are the same as those in \citep{}; )
have the same form as in \inftucker and are listed as follows:
\begin{align}
&q(z_{\bi}) \propto \N(\expec{q}{m_{\bi}},1)\delta(z_{\bi}>1),\label{eq:var_e_z}\\
&\expec{q}{z_{\bi}}=\expec{q}{m_{\bi}} + \frac{(2y_{\bi}-1)\N(\expec{q}{m_{\bi}}|0,1)}{\Phi((2y_{\bi}-1)\expec{q}{m_{\bi}})},
\end{align}
\begin{align}
&q(\vec(\Mcal_{nt})) = \N(\vec(\Mcal_{nt})|\bmu_{nt}, \bUpsilon_{nt}),\label{eq:var_e_M}\\
&\bmu_{nt} = \vec(\expec{q}{\Mcal_{nt}}) = \bUpsilon_{nt}~\vec(\expec{q}{\Zcal_{nt}}),  \\
&\bUpsilon_{nt} = \bLambda_{nt} \rbr{\I + \bLambda_{nt}}^{-1}.
\end{align}
where $\bi=\{i_1, \ldots, i_K\}$ indexes an entry in $\Zcal_{nt}$ and $\bLambda_{nt} = \bSigma_{nt}^{(1)} \otimes \ldots \otimes \bSigma_{nt}^{(K)}$ where $\bSigma_{nt}^{(j)} = k(\tU_{nt}^{(j)}, \tU_{nt}^{(j)})$ is the $j$-th mode covariance matrix over the sub-factors of $\tUcal_n$ corresponding to the subarray $\Ycal_{nt}$.
}

\subsection{Estimating latent factors}
Given the variational distributions, we estimate the group-specific latent factors $\{\tUcal_n\}_{n=1}^N$ and the common latent factors $\Ucal$ by maximizing the expected log joint probability,
\begin{align}
\expec{q}{\log p(\Ucal, \{\tUcal_n, \Ycal_n, \Zcal_n, \Mcal_n\}_{n=1}^N)}.
\end{align}
Specifically, we optimize the group-specific latent factors $\{\tUcal_n\}_{n=1}^N$ via SGD in the \map step and update the common latent factors $\Ucal$ in the \reduce step.

\subsubsection{Estimating the group-specific latent factors
$\{\tUcal_n\}$ via MAPPER}
Given $\Ucal$, the expected log likelihood function with respect to $\tUcal_n$ is
\begin{align}
f(\tUcal_n) &= \log(p(\tUcal_n|\Ucal)) \notag \\
 &+ \sum_{t=1}^{T_n} \big( \expec{q}{p(\Zcal_{nt}|\Mcal_{nt})} +  \expec{q}{\log(p(\Zcal_{nt}|\Mcal_{nt}))} \notag \\
 &+ \expec{q}{\log(p(\Mcal_{nt}|\tUcal_n))}\big). \label{eq:local_u_obj}
\end{align}
We have investigated L-BFGS to maximize Equation \eqref{eq:local_u_obj} over $\tU_n^{k}$. It turns out that SGD leads to better performance for our problem here.

To perform SGD, we first rearrange the objective function in Equation \eqref{eq:local_u_obj} as a summation form,
\begin{align}
f_n(\tUcal_n) &= \sum_{t=1}^{T_n} g_{nt}(\tUcal_n) \notag \\
g_{nt}(\tUcal_n) &= \frac{1}{T_n} \log(p(\tUcal_n|\Ucal)) + \expec{q}{p(\Zcal_{nt}|\Mcal_{nt})} \notag \\
            &+  \expec{q}{\log(p(\Zcal_{nt}|\Mcal_{nt}))} \notag \\
            &= -\frac{1}{2T_n\lambda} \sum_{j=1}^K \|\vec(\U^{(j)}) - \vec(\tU_n^{(j)})\|^2 \notag \\
            &+ \|\tucker{\expec{q}{\Mcal_{nt}}; (\bSigma_{nt}^{(1)})^{-\frac{1}{2}},\ldots,(\bSigma_{nt}^{(K)})^{-\frac{1}{2}}}\|^2 \nonumber\\
            &+ \sum_{k=1}^K \frac{m_{nt}}{m_{nt,k}}\log|\bSigma_{nt}^{(k)}| +  \tr\rbr{{\bLambda_{nt}}^{-1}\bUpsilon_{nt}}\label{eq:sgd_term}
\end{align}
where $m_{nt,k}$ is the dimension of $k$-th mode in $\Ycal_{nt}$, $m_{nt}=\prod_{k=1}^K m_{nt,k}$,
$\bLambda_{nt} = \bSigma_{nt}^{(1)} \otimes \ldots \otimes \bSigma_{nt}^{(K)}$, $\bSigma_{nt}^{(k)} = k(\tU_{nt}^{(k)}, \tU_{nt}^{(k)})$ is the $k$-th mode covariance matrix over the sub-factors of $\tUcal_n$, and $\bUpsilon_{nt}$ is the statistics computed in the variational E-step.

We randomly shuffle the subarrays in $\Ycal_n$ and sequentially process each subarray. For each subarray $\Ycal_{nt}$, we have the following update:
\begin{align}
\tUcal_n = \tUcal_n + \eta \partial g_{nt}(\tUcal_n). \label{eq:var_m_tu}
\end{align}
The gradient $\partial g_{nt}(\tUcal_n)$ has a form similar to that of the expected log joint probability with respect to global latent factors $\Ucal$ in InfTucker. We omit the detailed equation here and refer the detail to the paper by \citep{XuYQ12}.
The SGD algorithm is summarized in Algorithm \ref{alg:sgd}.  The SGD optimization for each $\tUcal_n$ is implemented by a \map task in the \mapreduce system.

\subsubsection{Estimating the parent latent factors
$\Ucal$ via REDUCER}
Given $\{\tUcal_1, \ldots, \tUcal_N\}$,
the expected log joint probability as a function of $\Ucal$ is
\begin{align}
f(\Ucal)
&=\sum_{n=1}^N \sum_{k=1}^K \log \N(\tU_n^{(k)}|\U^{(k)}, \lambda\I). \label{eq:var_m_u}
\end{align}
Setting the gradient of \eqref{eq:var_m_u} to zero, we have the simple update for $\Ucal$
\begin{align}
\U^{(k)} = \frac{1}{N}\tU_n^{(k)}. \label{eq:var_m_update_u}
\end{align}
We implement this step in the \reduce step of \mapreduce.
The algorithm is summarized in Algorithm 2.

\subsection{Algorithm complexity}
\cmt{
In this section, we present a distributed inference procedure of \ours over \mapreduce framework. Since updating   $\{\tUcal_n\}_{n=1}^N$ are independent given $\Ucal$, we can issue $N$ \mapper tasks, where each \mapper process a group of subarrays and perform SGD to update the corresponding $\tUcal_n$.
The estimating of $\Ucal$ needs to aggregate all the group-specific latent factors and we can assign a \reduce task correspondingly. The whole procedure is listed in Algorithm \ref{alg:dintucker}.
}
The time complexity of \InfTucker is $O(\sum_{k=1}^K m_k^3 + m_km)$ where $m_k$ is the dimension of the $k$-th mode and $m=\prod_{k=1}^K m_k$. If any $m_k$ is large, then \InfTucker is computationally too expensive to be practical.
For \ours, if the dimension of a subarray in mode $k$ is $\mhat_k$, the time complexity of analyzing it is $O(\sum_{k=1}^K \mhat_k^3 + \mhat_k\mhat)$ where $\mhat = \prod_{k=1}^K\mhat_k$ is the total number of entries in a subarray. When we set identical $\mhat_k$ for any $k$, the time complexity becomes $O(\mhat^{(1+\frac{1}{K})})$. Given $L$ subarrays and $N$ \mapper nodes, the time complexity for each \mapper node is $O(\frac{L}{N}\mhat^{(1+\frac{1}{K})})$, nearly linear in the number of elements in each small subarray.

The space complexity of \InfTucker is $O(m + \sum_{k=1}^K m_k^2)$ because it needs to store the whole array and the covariance matrices for all modes  in the memory of a computer. This is obviously infeasible for large data. By contrast, \ours only needs to store one small subarray and its covariance matrices 
in each \mapper node via streaming, and
the space complexity is $O(\mhat + \sum_{k=1}^K \mhat_k^2)$ where $r_k$ is the number of latent factors in mode $k$.
\alanc{to be fix.}

\cmt{
By choosing $\mhat_k$ to be small enough so that $\mhat \ll m$, the \mapper and \reducer nodes can store all the necessary information in main memory and perform the computation much faster than \InfTucker. Thus, \ours is feasible for inference on large multidimenstional arrays.
}

\begin{algorithm}                      
\caption{VB-SGD($\Ycal_n, T_n, \eta, \lambda$, $\Ucal$)}          
\label{alg:sgd}                           
\begin{algorithmic}                    
    \STATE Random shuffle subarrays in $\Ycal_n$.
    \STATE Initialize $\tUcal_n$ with $\Ucal$.
    \FOR {$t$=1 {\bfseries to} $T_n$}
        \STATE Pick up $t$-th subarray $\Ycal_{nt}$ in $\Ycal_n$
        \STATE Carry out variational E-step to optimize $q(\Mcal_{nt})$ and $q(\Zcal_{nt})$
        \STATE Calculate $\partial g_{nt}(\tUcal_n)$ and update $\tUcal_n$ according to Equation \eqref{eq:var_m_tu}.
    \ENDFOR
    \STATE {\bfseries return} $\tUcal_n$
\end{algorithmic}
\end{algorithm}

\begin{algorithm}
\caption{\ours($\{\Ycal_1, \ldots, \Ycal_N\}$, $\Ucal_0$, $T$, $\eta$, $\lambda$, $R$)}
\label{alg:dintucker}
\begin{algorithmic}
\STATE Initialize $\Ucal$ with $\Ucal_0$.
\REPEAT
    \FORALL  {$n \in \{1,\ldots, N\}$ \textbf{parallel} }
        \STATE \map task $n$: $\tUcal_n=\textrm{VB-SGD}(\Ycal_i, T, \eta, \lambda, \Ucal)$
    \ENDFOR
    \STATE \reduce task: Aggregate from all \map results $\{\tUcal_1, \ldots, \tUcal_N\}$ to update $\Ucal$, according to Equation \eqref{eq:var_m_update_u}.
\UNTIL {$R$ iterations}
\STATE {\bfseries return} $\Ucal$
\end{algorithmic}
\end{algorithm}

\subsection{Strategies for sampling subarrays}\label{sec::sampling_strategy}
\cmt{Here we discuss three ways to generate subarrays used in our training. These subarrays should have similar sizes so that the work load is balanced across \mapper nodes and there is no much waiting time on any node.}
Here we discuss three ways to generate subarrays used in our training. To optimize the performance of \mapreduce, we make these subarrays in the same size to ensure that the work load is balanced across MAPPER nodes.
To achieve this, we investigated three strategies.

\cmt{
To avoid waiting time between
Our model allows the flexibility of sampling subarrays. We can categorize the methods to generate a subarray into two types: we first sample the locations of factors in each mode, and then extract the subarray based on all the locations; Or we first sample the (non-zero) entries, and then extract a (minimum) subarray containing these entries. While the second type methods directly operate on the array entries, it is not trivial to control the subarray size. For example, two subarrays containing the same number of nonzero entries might have largely different sizes. Since the space and time complexity of our algorithm are sensitive to the subarray size, we need a strategy to easily control the subarray size so as to maintain a balanced work load in each computational node. For example, we may require all the subarrays to be of the same size and allocate identical number of subarrays on each \mapper node. Therefore, we focus on the first type methods, where we can easily control the subarray size. We devise three subarray-sampling strategies, described as follows.
}

\textbf{Uniform sampling.} This is the simplest method: we just uniformly sample a set of indexes of size $\mhat_k$, for each mode $k$, 
to define a subarray. To make multiple subarrays, we just repeat this process so that each subarray has the same size.

\textbf{Weighted sampling.} This strategy aims to let each subarray contain roughly the same number of nonzero elements (so that no subarray contains all zeros). In other words, we sample each nonzero element with the equal chance. This strategy is the same as the first one but with a critical difference: instead of sampling
a set of indexes uniformly for each mode, we sample these indexes based on weights of the corresponding array slices. The weight of an array slice is defined as the number of nonzero elements in the slice. Due to the weighted sampling, the numbers of nonzero elements in different subarrays are similar to each other.  A slice with a large weight contains rich information; for example, for the two-dimensional case, a slice corresponds to a network node and the large weight means that this node has many connections to other nodes. The weighted sampling strategy naturally gives more weights to these important slices (nodes).

\comment{
 Thus, the informative factors, which appear more often in the observed array and thus help better to capture the underlying structures, are more likely to appear in the sampled subarrays.
}

\textbf{Grid sampling.} It ensures the coverage of every element of the whole array. Specifically, we first randomly permute the indexes in each mode, then partition the permuted indexes into multiple segments with the same size, and repeat this process for each mode to generate a grid. In this grid, each (hyper-)cube contains a subarray. We can repeat this whole process to generate more subarrays.

\subsection{Predicting array entries by bagging}
\cmt{
To predict the value for a new array $\bi=\{i_1, \ldots, i_K\}$, we first sample a set of latent factors in each mode. Then we add the $i_1$-th latent factor in $\U^{(1)}$, the $i_2$-th latent factor in $\U^{(2)}$ and so on until all the latent factors corresponding to entry $\bi$ is added into the factors set. We denote the latent factor set by $\Ucal_\xi=\{\U_\xi^{(1)},\ldots,\U_\xi^{(K)}\}$. Based on the locations of factors in $\Ucal_\xi$, we can extract the subarray $\Ycal_{\xi}$. Then given $\Ucal_\xi$, we perform the variational E-step
 to obtain the posterior distribution of $q(\Mcal_\xi)$. We use the predictive posterior mean of $q(\Mcal_\xi)$ as the predicted value; that is, we use the value of the entry in $\expec{q}{\Mcal_\xi}$ corresponding to the same latent factors as $\bi$ as the prediction result. We can repeat the procedure for several times and take the average prediction value as the final prediction. For a batch of test entries, we can predict their values in the same way.

Note that similar to the learning procedure, our procedure for predicting a new array entry is  based on small subarrays containing this entry. \InfTucker, by contrast, needs to infer the whole latent tensor $q(\Mcal)$ in order to predict the specific entry value, which is infeasible for large arrays.
}

To predict the values of unknown entries, the original \InfTucker needs to infer the posterior distribution of the whole latent array. For large arrays, this inference is computationally prohibitive. To overcome this hurdle, we apply a bagging strategy which  learns the prediction by simply aggregating predictions on a collection of small subarrays. Because \ours can quickly provide predictions on the small subarrays, it achieves fast final predictions.
Note that Bagging \citep{trevor2001elements} has been widely used to improve prediction accuracy for many machine learning methods such as neural networks and decision trees. For \ours, we first
generate subarrays and find their corresponding latent factors, then use them to learn predictive means of the unknown elements following the GP prediction algorithm in \InfTucker (but on the subsets here), and finally aggregate the predictive means by averaging. As we sample subarrays from the whole array, our prediction can be viewed as nonparametric bootstrap prediction \citep{fushiki2005nonparametric}.

\section{Related work}
Our work is naturally built upon InfTucker \citep{XuYQ12} and is closely related to
the random function prior model \citep{LloydOGR12randomgraph}, a generalization of InfTucker. \ours scales up the inference of  InfTucker on massive multidimensional array data based on the hierarchical Bayesian treatment and  enables local computation via the Mapper function and global information sharing via the Reducer function. This divide-and-conquer strategy is general and can be used to train other special instances of the random function model such as
the infinite relational models \citep{Kemp06learningsystems} and GP latent variable models (GP-LVMs) \citep{Lawrence06GPLVM} on large data.

Actually our strategy can also be applied to train classical tensor decomposition models, as an alternative to GigaTensor. On one hand, one advantage of using our approach over GigaTensor is that we can easily control the computational cost by tuning the number and sizes of subarrays (with the trade-off between speed and accuracy).
We can also readily conduct either Tucker or PARAFAC decomposition based on our strategy, while GigaTensor is currently limited to PARAFAC. On the other hand, to speed up the computation, GigaTensor exploits sparsity in data while our approach does not. For  applications where the multidimensional arrays are dense such as fMRI data, our approach is well suited. But for applications where the arrays are sparse such as NELL data used in our experiment, then exploiting sparsity as GigaTensor can further  speed up our distributed inference (note that even without utilizing sparsity in data, \ours is faster than GigaTensor with higher prediction accuracy.)

\cmt{
 Both \InfTucker and \ours are nonparametic models based on Gaussian process.\cmt{ While \InfTucker assumes the whole tensor is projection of the tensor-variate Gaussian process over a the latent factor matrices $\Ucal=\{\U^{(1},\ldots, \U^{(K)}\}$, \ours assumes the tensor entries come from a series of sub-tensors and each sub-tensor is generated by the projection of the tensor-variate Gaussian process over the corresponding latent factors.} Another nonparametric tensor decomposition model is Random Function Prior model (RFP) prosed by \citet{zoubin}. Based on the De Finetti-type
 presentations for random arrays, the RFP model assumes any set of tensor entries are projection a Gaussian process.
 }

\section{Experiment}
To evaluate \ours, we performed experiments to answer the following questions:

\noindent \textbf{Q1} How does the distributed online inference of \ours compare to the sequential inference of InfTucker?\\
\textbf{Q2} How does \ours scale with regard to the number of machines?\\
\textbf{Q3} How does \ours perform on real-world multidimensional arrays with billions of entries and compare with GigaTensor, the state-of-the-art
tensor decomposition method,  in terms of both prediction accuracy and running time?

To answer the first question, we examined \ours on three small datasets for which InfTucker is computationally feasible, as described in Section \ref{sec:exp-small}. To answer the second and third questions, we used two large real datasets in Sections \alanc{Section?} 6.2 and 6.3.

We carried out our experiments on a \hadoop cluster. The cluster consists of 16 machines,  each of which has a 4-quad Intel Xeon-E3 3.3 GHz CPU, 8 GB RAM, and a 4 Terabyes disk. We implemented \ours with PYTHON and used \hadoop streaming for training and prediction.

\begin{figure*}[t!]
\centering
\subfigure{
\raisebox{12ex}{
\includegraphics[scale=0.32]{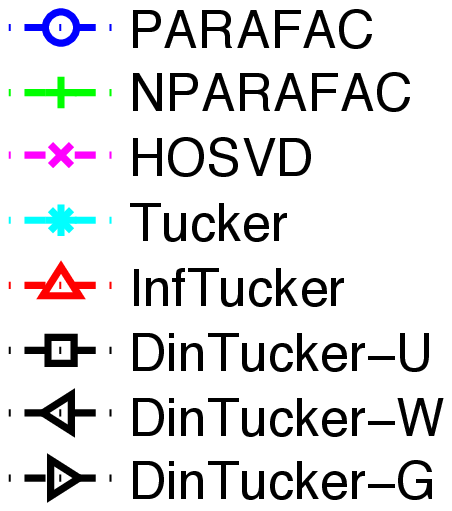}
}
}
\setcounter{subfigure}{0}
\hspace{-0.2in}
\subfigure[\textit{digg1}]{
\includegraphics[width=1.9in,height=1.5in]{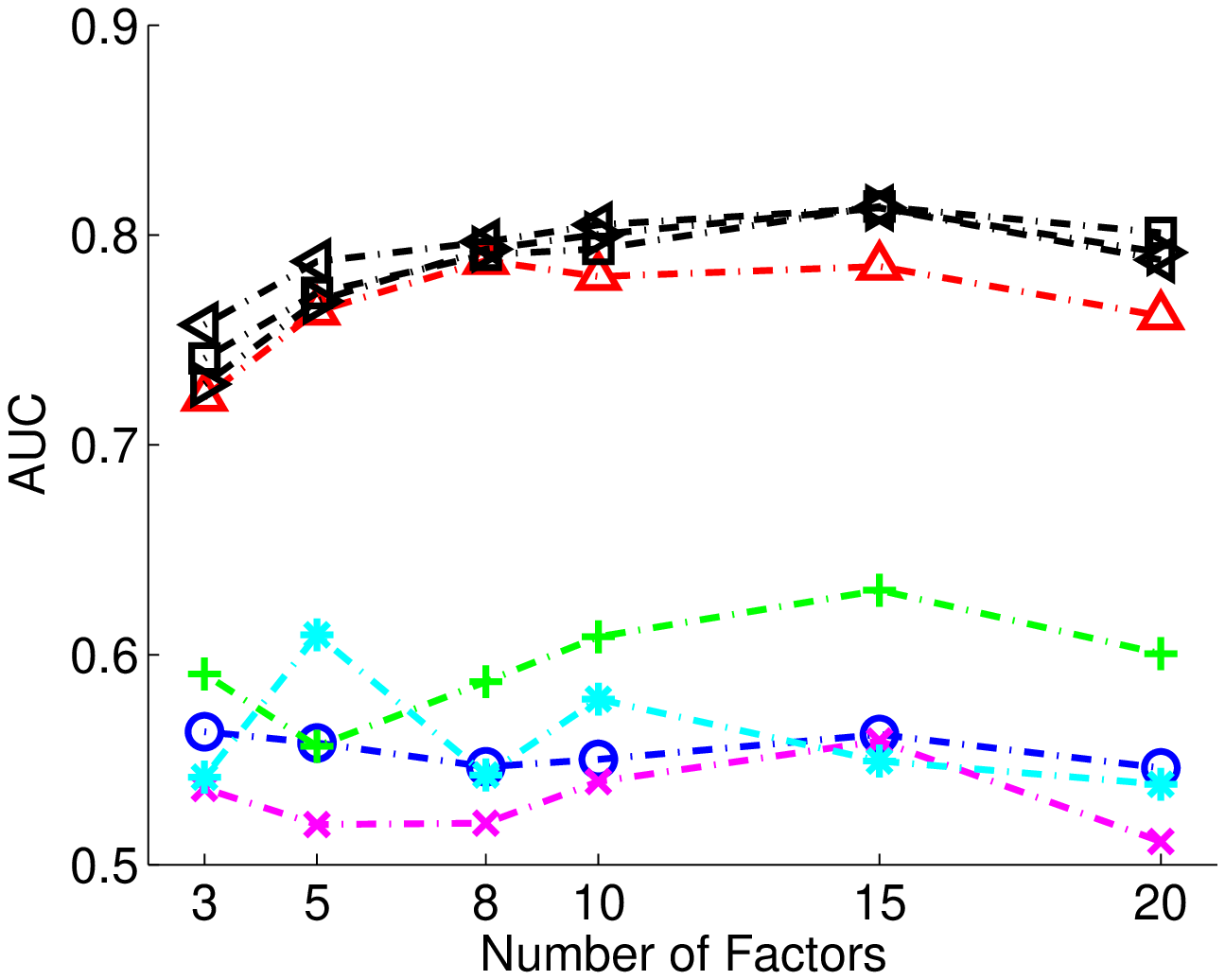}
}
\subfigure[\textit{digg2}]{
\includegraphics[width=1.9in,height=1.5in]{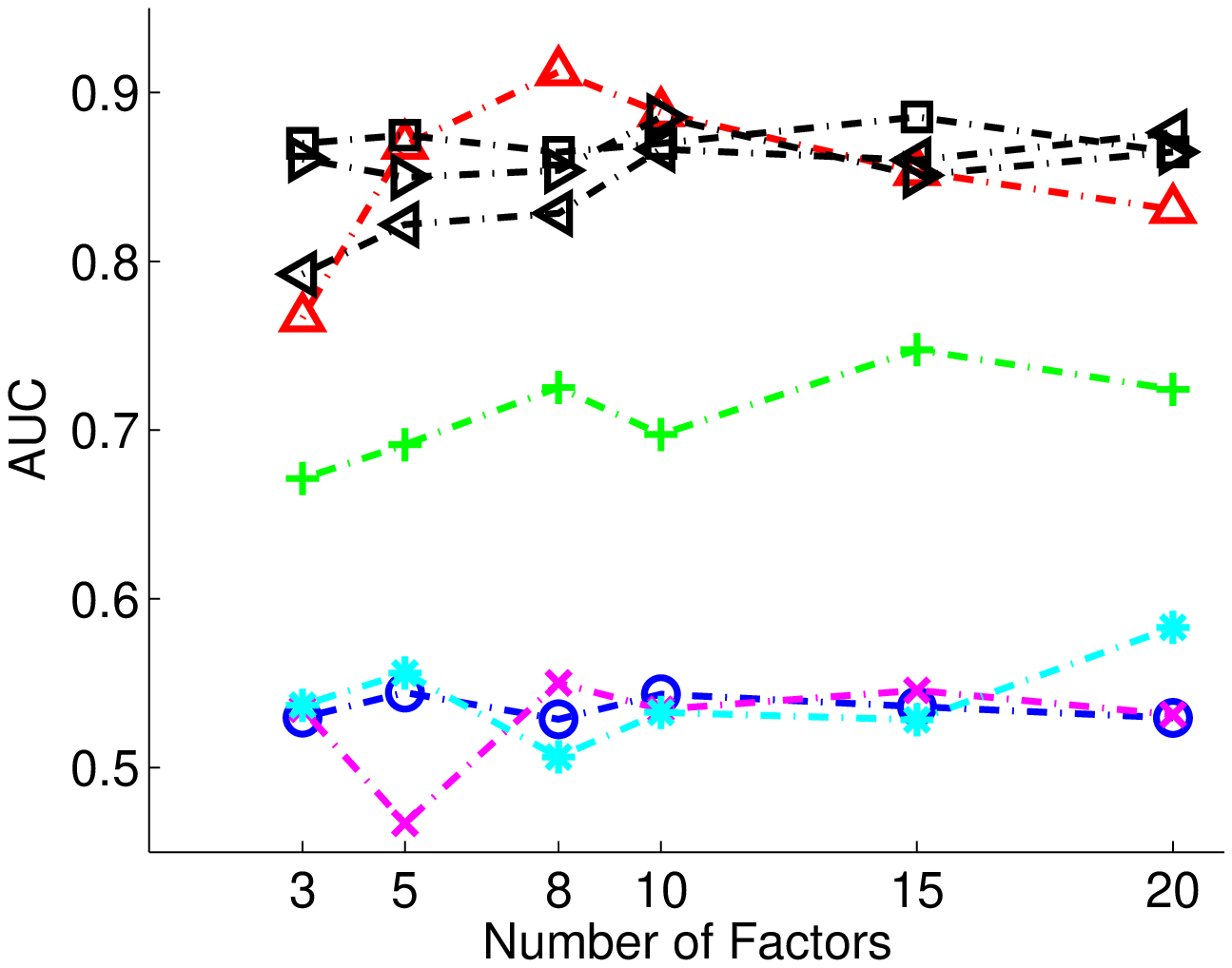}
}
\subfigure[\textit{enron}]{
\includegraphics[width=1.9in,height=1.5in]{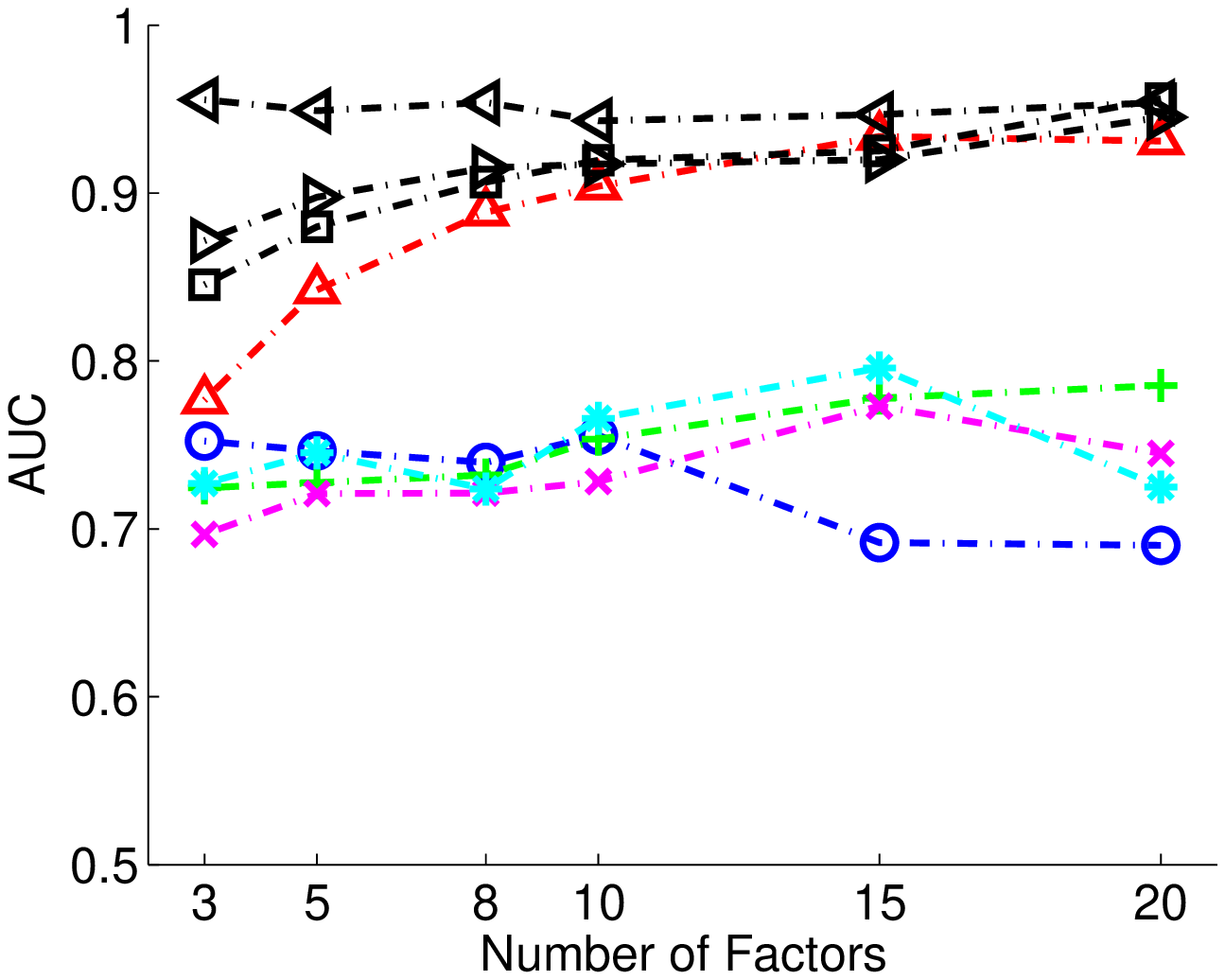}
}

\caption{The prediction results on small datasets. The results are averaged over 5 runs. \oursu, \oursw and \oursg refer to our method based on the uniform, weighted, and grid sampling strategies, respectively.}
\label{fig:small-auc}
\end{figure*}

\subsection{Small datasets}\label{sec:exp-small}
We first examined \ours on the following social network datasets, \textit{Digg1}, \textit{Digg2} and \textit{Enron}. Both \textit{Digg1} and \textit{Digg2} datasets are extracted from a social news website \url{digg.com}. \textit{Digg1} describes a three-way interaction (news, keyword ,topic), and \textit{Digg2} a four-way interaction (user, news, keyword, topic). \textit{Digg1} contains $581 \times 124 \times 48$ elements and 0.024\% of them are non-zero. \textit{Digg2} has $22 \times 109 \times 330 \times 30$ elements and 0.002\% of them are non-zero. \textit{Enron} is extracted from the Enron email dataset. It depicts a three-way relationship (sender, receiver, time). The dataset contains $203 \times 203 \times 200$ entries, of which 0.01\% are nonzero.

We compared \ours with the following tensor decomposition methods: PA-\\RAFAC, nonnegative PARAFAC (NPARAFAC)\citep{ShashuaH05}, high order SVD (HOSVD) \citep{Lathauwer00HOSVD}, Tucker decomposition and \InfTucker. We chose the number of latent factors from the range \{3,5,8,10,15,20\}. Since the data are binary, we evaluated all the approaches by the area-under-curve (AUC) based on a random 5-fold partition of the data. Specifically, we split the nonzero entries into $5$ folds and used $4$ folds for training. For the test set, we used all the ones in the remaining fold and randomly chose 0.1\% zero entries (so that the evaluations will not be overwhelmed by zero elements). We repeated this procedure for 5 times with different training and test sets each time. For \InfTucker, we used cross validation to tune the hyperparameter of its Laplace prior. For \ours, we set the subarray size to  $40 \times 40 \times 40$ for \textit{Digg1} and \textit{Enron}, and $20 \times 20 \times 20 \times 20$ for \textit{Digg2}. We used the three strategies described in Section \ref{sec::sampling_strategy}. To generate subarrays for training, for each strategy, we sampled $1,500$ subarrays. We ran our distributed online inference algorithm with 3 mappers, and set the number of iterations to 5.
\cmt{
For each dataset, we sample $500$ subarrays. To sample a subarray,  we randomly sample indices for each mode and then use the index sets to extract the  corresponding subarray. }
We tuned the learning rate $\eta$ in Equation \eqref{eq:var_m_tu} from the range \{0.0005, 0.001, 0.002, 0.005, 0.01\}. We used another cross-validation to choose the kernel function from the RBF, linear, Polynomial and Mat\'ern functions and tuned its hyperparameters. For the Mat\'ern kernel, the order of its Bessel function is either $\frac{3}{2}$ or $\frac{5}{2}$. For our bagging prediction, we randomly sampled 10 subarrays, each with the same size as the training subarrays.\cmt{, and predict a batch of 10 entries at one time. We simply use the uniform sampling strategy to generate these subarrays for prediction.} The results are shown in Figure \ref{fig:small-auc}. As we can see, in terms of the AUC accuracy, all versions of \ours are similar to InfTucker on \textit{Digg2} and better than InfTucker on \textit{Digg1} and \textit{Enron}. Furthermore, \ours significantly outperforms all the other alternative methods.

\subsection{Scalability with regard to the number
of machines}
To examine the scalability and predictive performance of \ours, we used the following large datasets in two real-world applications.
\begin{itemize}
\item NELL: Knowledge bases containing triples (\eg 'George Harrison', 'playsInstrument', 'Guitar') from the 'Read the Web' project \citep{carlson2010toward}. This dataset is downloaded from \url{http://rtw.ml.cmu.edu/rtw/resources}. We filtered out the triples with confidence less than 0.99 and then analyzed the triplets from 20,000 most frequent entities.
\item ACC: Access logs from a source code version control system in a large company. The log provides various information such as user id, target resource (\ie file name), action (\ie "FileCheckIn" and "FileCheckOut"), the start time and end time of the action. 
We used the records from 2000 most active users and extracted triples (user, action, resource) for analysis.
\end{itemize}
The statistics of the datasets are summarized in Table \ref{tb:large_data}.
\begin{table}[htbp]
\caption{Statistics of multidimensional array data. B: billion, K: thousand.}
\centering
\begin{tabular}{ccccc}
\hline\hline
{Data} & {I} & {J} & {K} & {Number of entries}\\
\hline
NELL & 20K & 12.3K & 280 & 68.9B\\
ACC & 2K & 179 & 199.8K & 71.5B\\
\hline\hline
\end{tabular}\label{tb:large_data}
\end{table}
\begin{figure}[ht]
\centering
\includegraphics[scale=0.35]{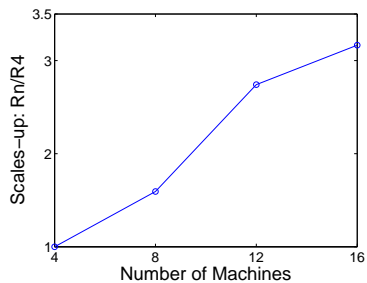}
\caption{The scalability of \ours with regard to the number of machines on the NELL dataset. Note that the running time scales up linearly.}
\label{fig:scale}
\end{figure}
We examined the scalability of \ours with regard to the number of machines on the NELL dataset. We set the number of latent factors to 5 in each mode. We set the subarray size to $50 \times 50 \times 50$. We randomly sampled 590,400 subarrays, so that the number of array entries processed by \ours is roughly the same as the whole array: $50 \times 50 \times 50 \times 590400/(20000 \times 12295 \times 280) = 1.07$.
The results are shown in Figure \ref{fig:scale}. The Y-axis shows $R_n/R_4$, where $R_n$ is the running time for $N$ machines. Note that the running time scales up linearly.

\begin{figure*}[ht]
\centering
\subfigure[NELL: running time]{
\includegraphics[scale=0.32]{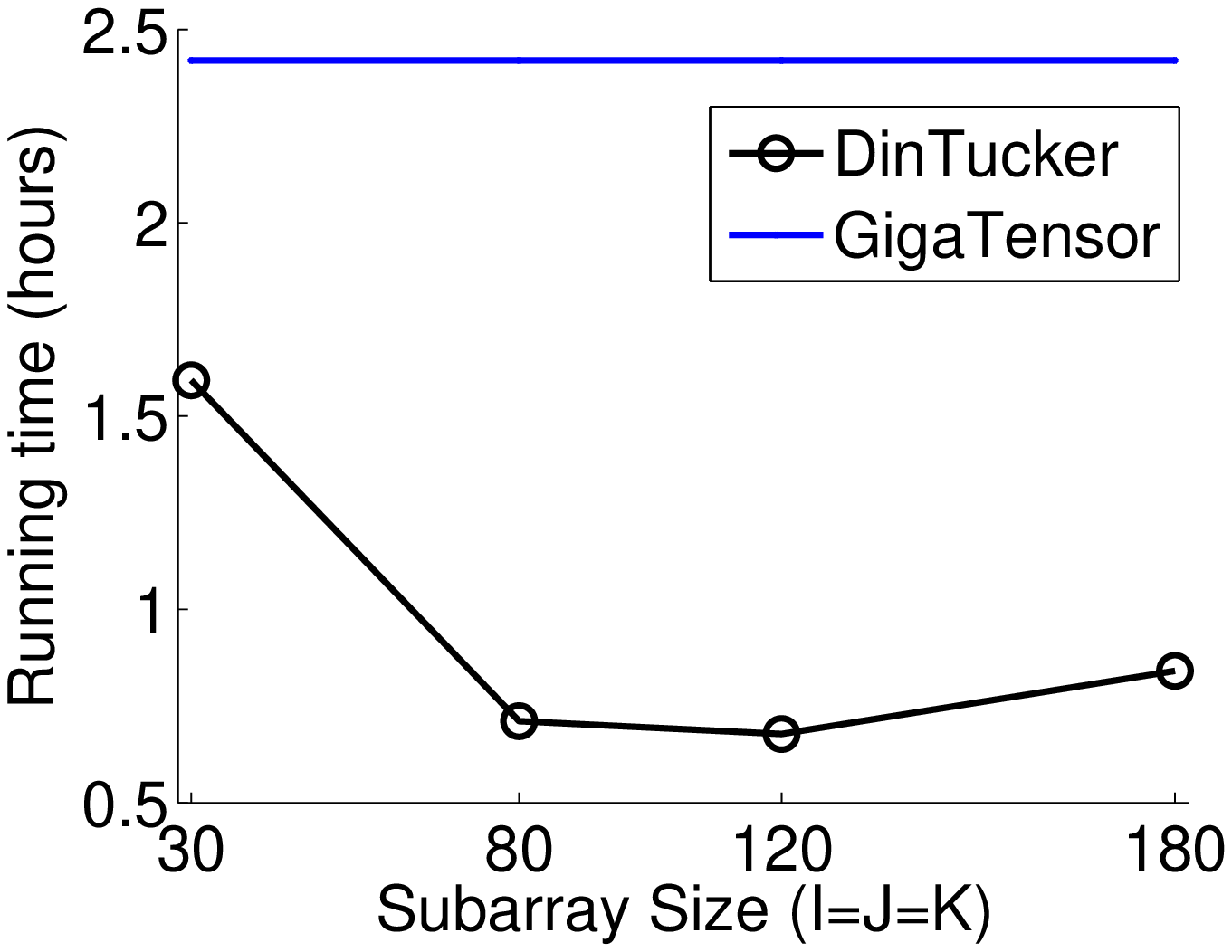}
}
\subfigure[NELL: prediction]{
\includegraphics[scale=0.32]{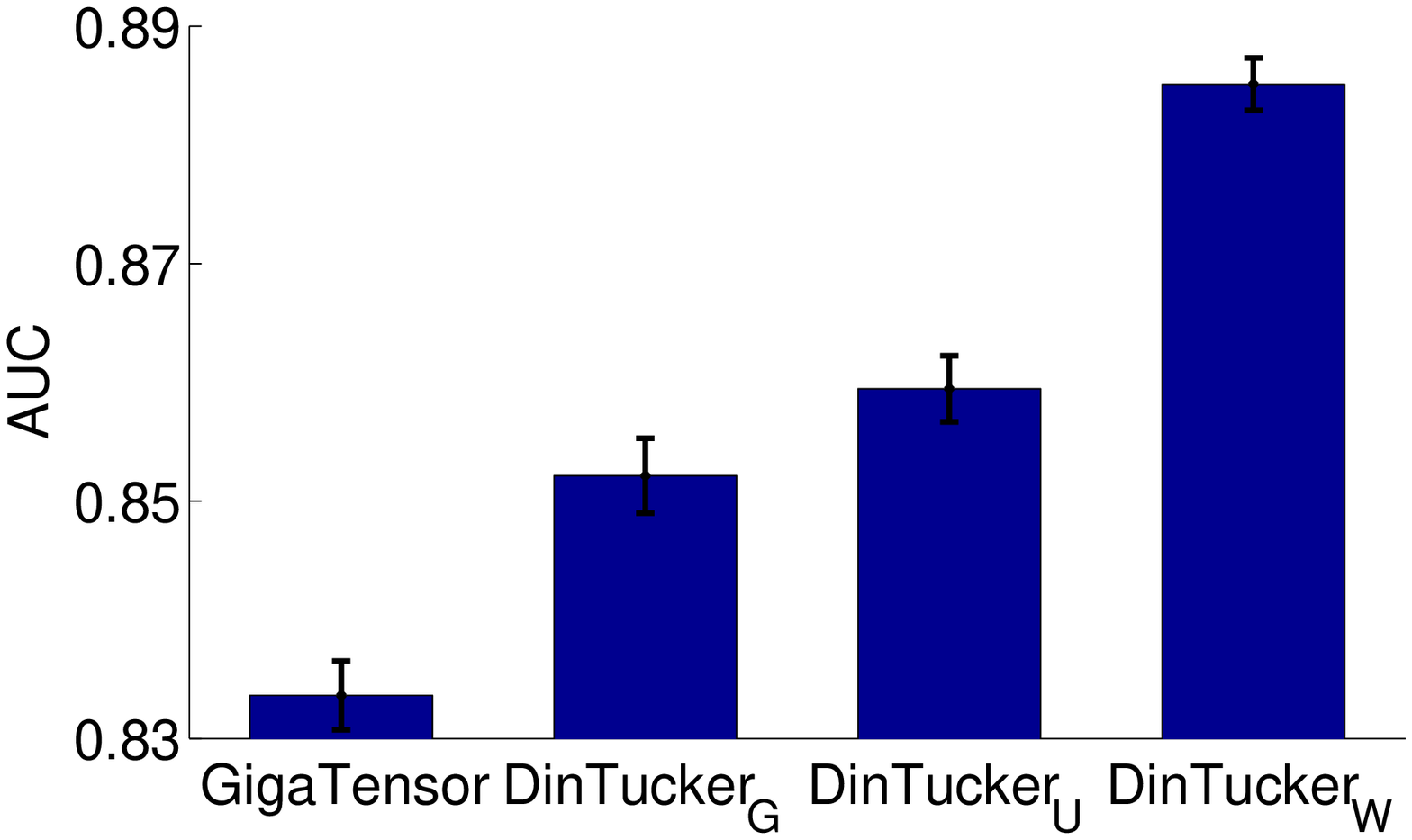}
}
\hspace{0.5in}
\subfigure[ACC: running time]{
\includegraphics[scale=0.32]{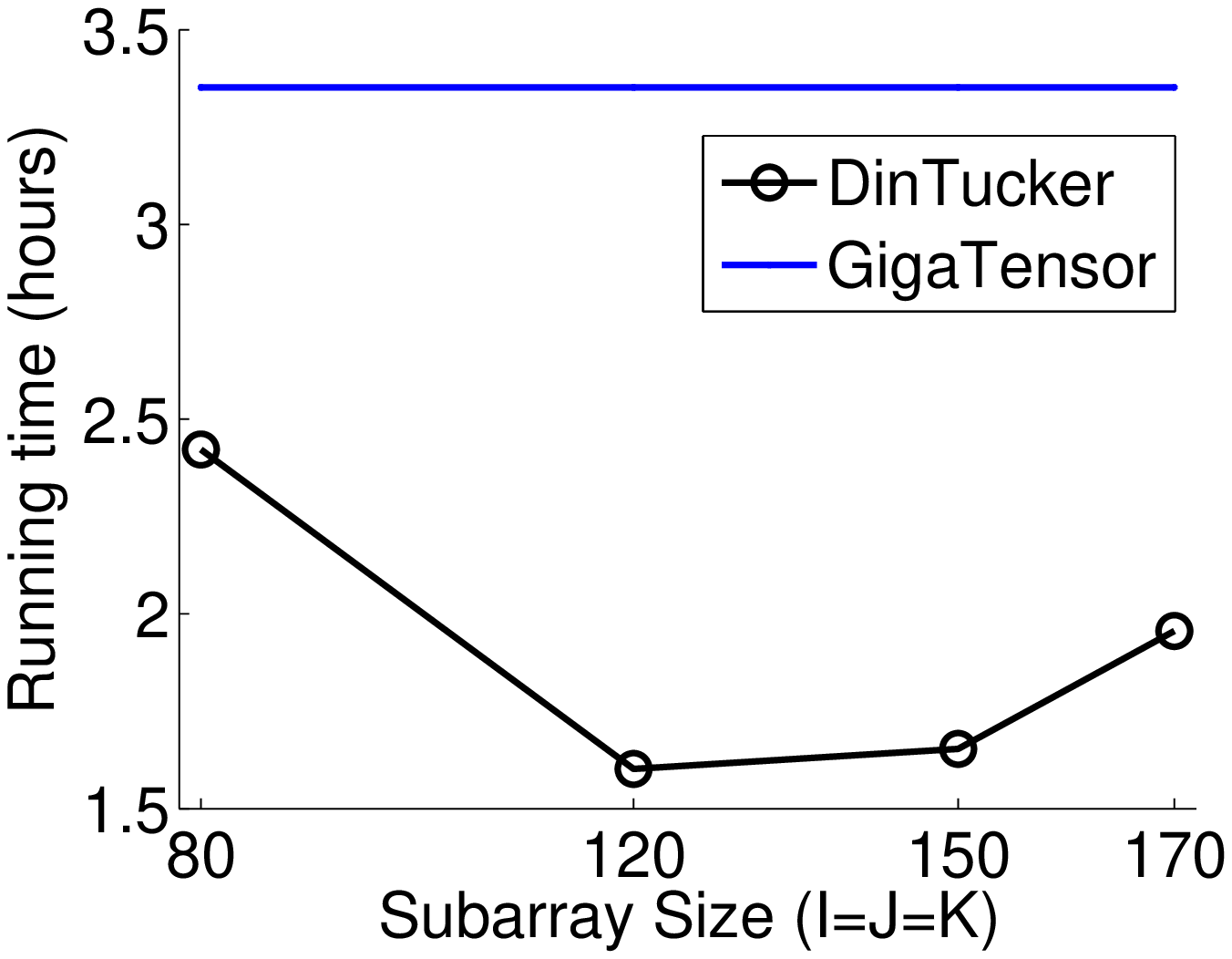}
}
\subfigure[ACC: prediction]{
\includegraphics[scale=0.32]{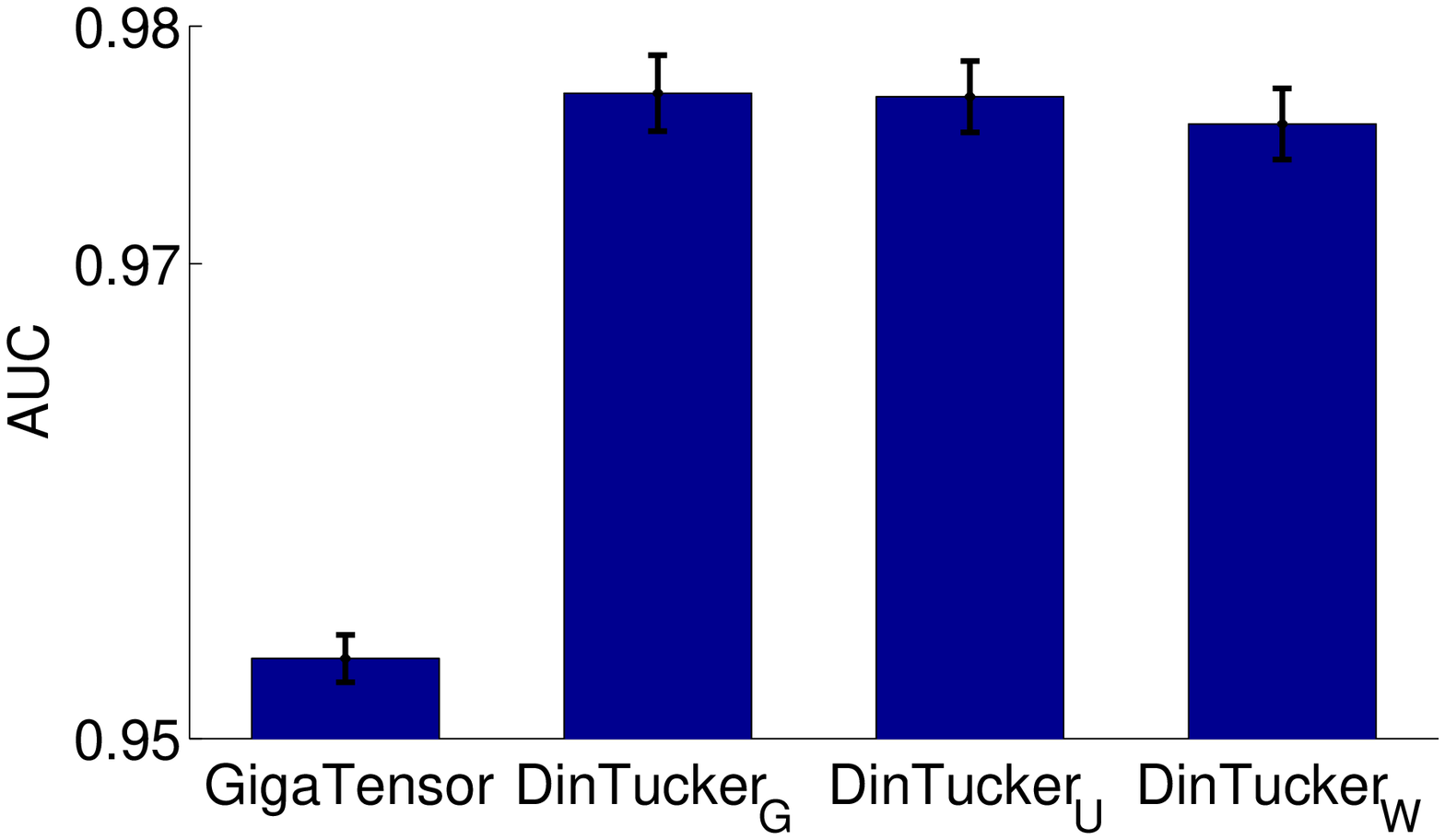}
}
\cmt{
\subfigure[NELL: prediction]{
\includegraphics[scale=0.33]{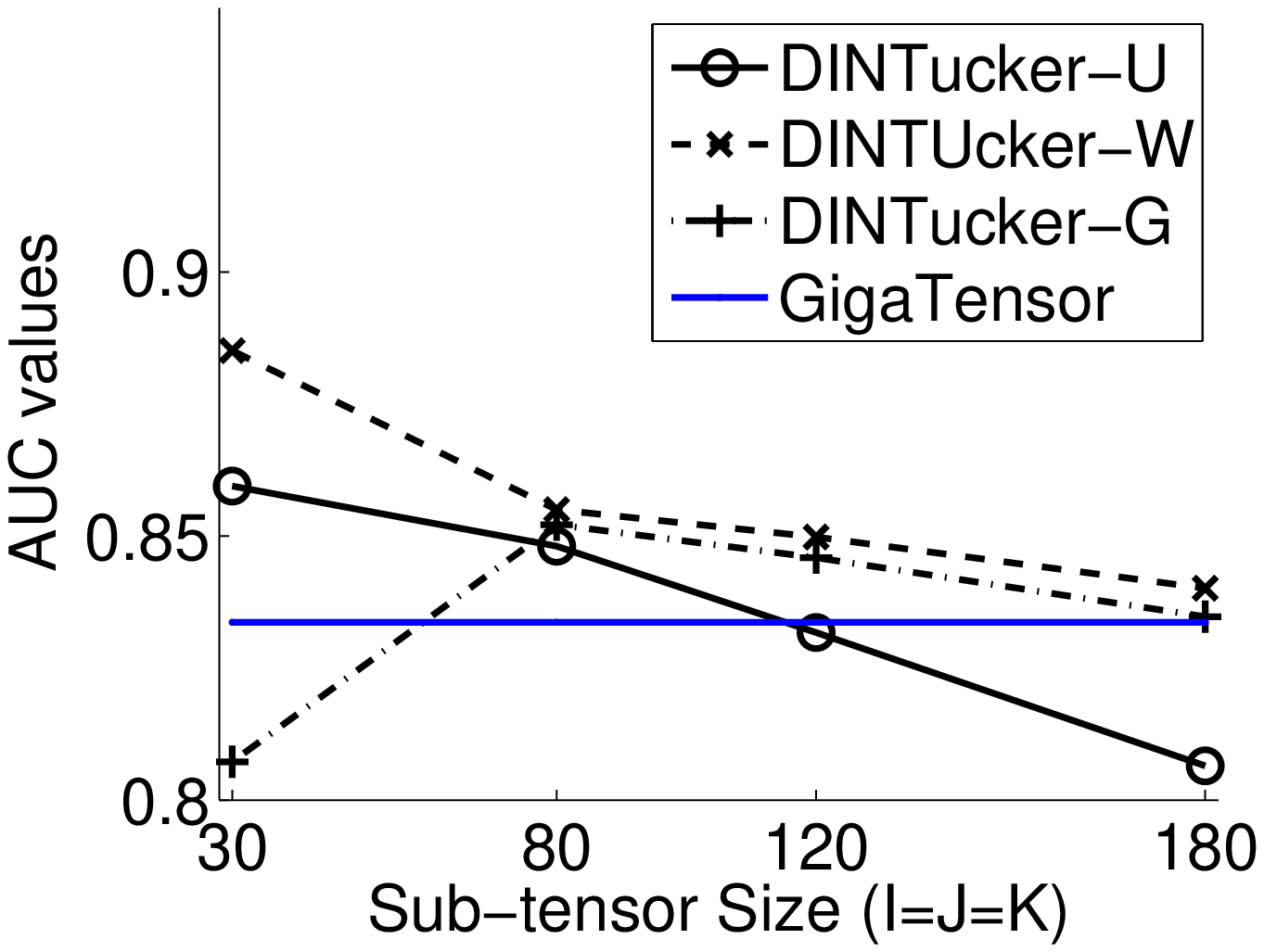}
}
\subfigure[ACC: prediction]{
\includegraphics[scale=0.33]{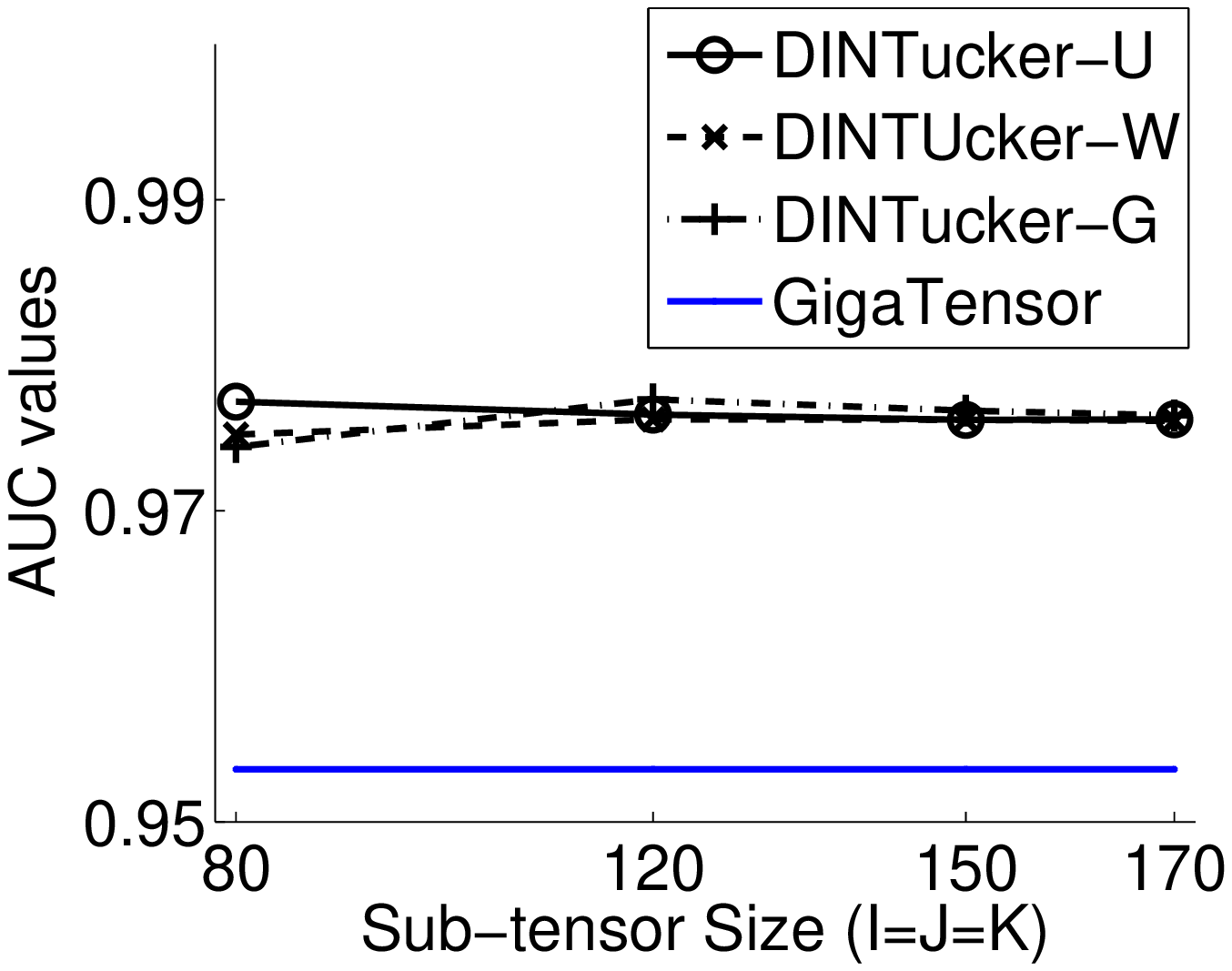}
}
}
\caption{The running time and AUC for the NELL and ACC datasets. The results are averaged over 50 test datasets.
}
\label{fig:runtimeauc}
\end{figure*}

\subsection{Running time and prediction accuracy}
We compared \ours with GigaTensor  on the NELL and ACC datasets.
\cmt{We run \ours and GigaTensor on a \hadoop cluster consisting of 16 machines where each machine has 4-quad Intel Xeon-E3 3.3 GHz CPU, 8 GB RAM and 4 Terabyes disk.} We used the original GigaTensor implementation in JAVA and adopted its default setting. 
For \ours, we set the \mapreduce iteration number to $5$ and used the Mat\'ern kernel. 

\comment{
\begin{table}[htbp]
\caption{The number of subarrays with different sizes. M: million, K: thousand.}
\centering
\subtable[NELL]{
\begin{tabular}{c|cccc}
\hline
{Size (I=J=K)} & 30 & 80 & 120 & 180 \\
{Number} & 2.55M & 134.48K & 39.84K & 11.81K\\
\hline
\end{tabular}
}
\subtable[ACC]{
\begin{tabular}{c|cccc}
\hline
{Size (I=J=K)} & 80 & 120 & 150 & 170 \\
{Number} & 139.73K & 41.40K & 21.20K & 14.56K\\
\hline
\end{tabular}
} \label{tb:numbervssize}
\end{table}
}

We set the number of latent factors for each mode to 5 for the NELL dataset and 10 for the ACC dataset.
\cmt{In the NELL dataset, the nonzero entries are 0.0001\% and, in the ACC dataset, the nonzero entries are 0.003\%.}
The NELL and ACC datasets contain 0.0001\% and 0.003\%  nonzero entries, respectively.
 We randomly chose 80\% of nonzero entries for training and then, from the remaining entries, we sampled $50$ test datasets, each of which consists of $200$ nonzero entries and $2,000$ zero entries. For {\ours}'s prediction, we randomly sampled 10 subarrays of size $50\times 50 \times 50$  for bagging.

To make a fair comparison, we trained \ours and GigaTensor using the same amount of data, which is the product of the sizes of the sampled subarrays and the number of the subarrays in the training. Also, to examine the trade-off between using fewer larger subarrays vs. using more smaller subarrays given the same computational cost, we varied the size of subarrays but kept the total number of entries for training to be the same as the number of entries in the whole array.

\cmt{;  used the same number of tensor entries
, we varied the number of subarrays to different sizes so that the total number of entries is the same as the original whole array.
For example, in NELL dataset, the number of subarrays is set to $2,550,075$ for size $30 \times 30 \times 30$ and $134,477$ for size $80 \times 80 \times 80$. \cmt{Then in each iteration, \ours process one-fifth of the subarrays. Thus after five iterations, \ours process all the subarrays, which contain the same number of entries as the whole array.}
The used subarray numbers and sizes are listed in Table \ref{tb:numbervssize}. Note that when we increase the size of subarray, we process less samples (thus lower the communication/IO cost), however, the time to process each subarray increases and vice versa.
}

Figure \ref{fig:runtimeauc} summarizes
the running time and AUC of \ours and GigaTensor on the NELL and ACC datasets.
The training time of \ours is given in Figures \ref{fig:runtimeauc}a and c. Note that since the training time only depends on the number and the size of subarrays, the three subarray sampling strategies described in Section \ref{sec::sampling_strategy} do no affect the training time.
Figures \ref{fig:runtimeauc}a and c also demonstrate the trade-off between the communication cost and the training time over the subarrays: if we use smaller subarrays, it is faster to train the GP model over each subarray, but it incurs a larger communication/IO cost.
As subarrays get smaller, the overall training time first decreases---due to less training time on each subarray---and then increases when the communication/IO cost is too large.
Figures \ref{fig:runtimeauc}b and d report the AUCs of GigaTensor and \ours based on different sampling strategies with subarray size $80\times 80 \times 80$.
They show that the weighted sampling strategy gives comparable or better results than the other methods, 
confirming the benefit of
giving larger sampling weights informative array slices (\ie ensuring that each nonzero element
has the equal chance to be used in the training).
Also, regardless the subarray sampling strategy, \ours outperforms GigaTensor consistently.
Although GigaTensor explores data sparsity for fast computation, \ours achieves more accurate prediction with faster training.


\section{Conclusion}

In this paper, we propose \ours, a nonparametric Bayesian learning algorithm that scales to large tensors. On small datasets, \ours achieves the same prediction accuracy as InfTucker. On large datasets for which InfTucker and other random function prior models are infeasible, \ours can train the model with ease.  Compared with the state-of-the-art distributed tensor decomposition method, GigaTensor,  \ours provides higher prediction accuracy 
and faster training speed.

\cmt{
 The main idea of {\ours}---hierarchical Bayesian modeling with stochastic gradient descent---can be used to train  other random function prior models (\eg GP-LVMs) as well as classical Tucker or PARAFAC decomposition models.
And it can also be readily implemented in other parallel computing platforms such as GPUs.
}

\bibliographystyle{newapa}
\bibliography{dintucker}
\end{document}